\pdfoutput=1

\documentclass[11pt]{article}

\usepackage[]{acl}
\usepackage{amsmath}
\usepackage{graphicx}
\usepackage{tikz}
\usepackage{times}

\definecolor{hotpink}{RGB}{255, 83, 115}
\definecolor{teal}{RGB}{90, 200, 250}
\definecolor{lightgreen}{RGB}{33, 222, 128}
\definecolor{lightblue}{RGB}{72, 123, 232}

\newcommand{\hide}[1]{}
\usepackage{booktabs}
\usepackage{tabularx}
\usepackage{balance}

\usepackage{times}
\usepackage{graphics}
\usepackage{multirow}
\usepackage{latexsym}

\usepackage[T1]{fontenc}

\usepackage[utf8]{inputenc}

\usepackage{microtype}

\title{Robustness of Fusion-based Multimodal Classifiers to Cross-Modal Content Dilutions}

\author{Gaurav Verma \\
        Georgia Institute of Technology \\
        \texttt{\small{gverma@gatech.edu}} 
        \And
        Vishwa Vinay \and Ryan A. Rossi \\
        Adobe Research\\
        \texttt{\small{\{vinay, ryrossi\}@adobe.com}} 
        \And
        Srijan Kumar \\
        Georgia Institute of Technology \\
        \texttt{\small{srijan@gatech.edu}}
        }        

\begin{document}
\newcommand{\mystar}{{\fontfamily{lmr}\selectfont$\star$}}
\maketitle
\begin{abstract}

As multimodal learning finds applications in a wide variety of high-stakes societal tasks, investigating their robustness becomes important. Existing work has focused on understanding the robustness of vision-and-language models to \textit{imperceptible} variations on benchmark tasks. In this work, we investigate the robustness of multimodal classifiers to \textit{cross-modal dilutions} -- a \textit{plausible} variation. We develop a model that, given a multimodal (image + text) input, generates additional dilution text that \textit{(a)} maintains relevance and topical coherence with the image and existing text, and \textit{(b)} when added to the original text, leads to misclassification of the multimodal input.  
Via experiments on Crisis Humanitarianism and Sentiment Detection tasks, we find that the performance of task-specific fusion-based multimodal classifiers drops by $23.3\%$ and $22.5\%$, respectively, in the presence of dilutions generated by our model. Metric-based comparisons with several baselines and human evaluations indicate that our dilutions show higher relevance and topical coherence, while simultaneously being more effective at demonstrating the brittleness of the multimodal classifiers. Our work aims to highlight and encourage further research on the robustness of deep multimodal models to realistic variations, especially in human-facing societal applications. 
\end{abstract}

\section{Introduction}
Rich multimodal content understanding is crucial for several AI for Social Good applications like humanitarian information detection during crises, hate speech analyses, and fake news mitigation~\cite{ofli2020analysis, kiela2020hateful, FacebookHatefulMemes, khattar2019mvae, verma2022overcoming}. In many such scenarios, the information in individual modalities, either image or text, is designed to be complementary to information in the other modality. As such, joint modeling of both modalities is of fundamental importance, and consequently, technologies that enable multimodal understanding are advancing rapidly and are being deployed at scale~\cite{GoogleMUM, grauman2021ego4d}. 

\begin{figure}
    \centering
    \includegraphics[width = \linewidth]{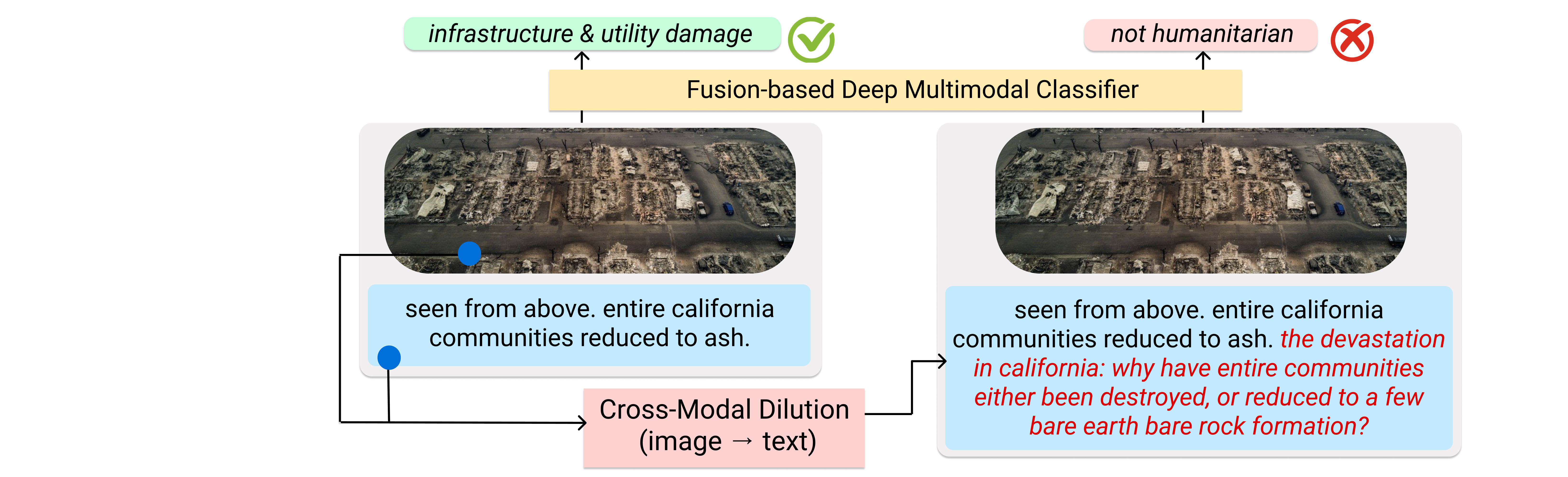}
    \caption{{\textbf{Overview of our study.} We investigate the robustness of fusion-based deep multimodal classifiers to cross-modal dilutions. We generate dilutions that maintain semantic relevance with the original text and image while causing incorrect classifications. We also demonstrate the realistic nature of cross-modal dilutions using human evaluation. The figure shows an actual example from our experiments.}}
    \label{fig:overview_figure}
    \vspace{-4mm}
\end{figure}

It is desirable that deep learning models are robust to dilution-based variations in input. \textit{Dilution} is defined as the addition of related content that dilutes the effect of the original information.
~\citeauthor{naik2018stress} (\citeyear{naik2018stress}) and \citeauthor{ribeiro2020beyond} (\citeyear{ribeiro2020beyond}) argue that natural language processing (NLP) models should not alter their predictions after adding dilutions --- for instance, appending statements like ``and true is true'' (multiple times) for the Natural Language Inference task and adding randomly created URLs for the Sentiment Analysis task. 

We study the robustness of multimodal classifiers to dilutions.
Compared to the simple dilutions created for NLP tasks, we aim to explore realistic dilutions for multimodal data. Since what entails plausible dilutions for multimodal data has not been established, we propose a new category of dilutions specific for multimodal content, named  \textit{cross-modal} dilutions. Cross-modal dilution involves adding relevant information from the image modality to the text modality for a multimodal input; see Figure \ref{fig:overview_figure}. Our notion of dilution, unlike the examples above, is contextual -- that is, the change introduced varies for different information items. Additionally, evaluating robustness to dilutions in a multimodal setting is non-trivial because the possible additions are constrained by the semantics of both the image and the original text.

Previous research on the robustness of deep multimodal learning focuses on perturbations for Visual Question Answering (VQA)~\cite{srivastava2020visual, zhang2019information, gupta2017survey, wu2017visual} and involves making minor alterations to the textual questions~\cite{mudrakarta2018did}, or asking more challenging questions than what were present in the training dataset~\cite{sheng2021human, li2021adversarial}. In contrast, we focus on multimodal classification and study dilution-based variations. To this end, we propose a method that leverages a large language model to generate \textit{additional} text that is \textit{(i)} related to the information in the image, \textit{(ii)} semantically aligned to the existing user-provided textual description, and \textit{(iii)} is adversarial in nature (i.e., when added to the existing description, leads to incorrect predictions by multimodal models). The first two constraints ensure that the additional text is realistic, while the third constraint enables us to assess the robustness of multimodal classifiers under these settings.

\noindent Our contributions are summarized as follows:\\
$\bullet$ We propose and investigate the robustness of multimodal classifiers to cross-modal dilutions. We develop an approach that leverages keywords from image and text to perform controlled generation of semantically relevant text that can be appended to the original text to cause misclassification. \\
$\bullet$ Via extensive evaluation covering aspects like adversarial effectiveness, content relevance, diversity, and coherence, we establish that the dilutions generated by our proposed model are better than several rule-based and model-based baselines. We release our code to aid future research.\footnote{Project webpage with code: \url{https://claws-lab.github.io/multimodal-robustness/}}\\
$\bullet$ We conduct human evaluations to \textit{(a)} assess the quality of generated dilutions over the most competitive baseline and \textit{(b)} establish the realistic nature of diluted multimodal examples. We find that our cross-modal dilutions are perceived by humans as better than the baseline dilutions and more realistic. 

\vspace{-1mm}
\section{Related Work}\label{sec:related-work}
\vspace{-1mm}
\textbf{Robustness of Multimodal Models}: \label{sec:related-robustness-multimodal}
Existing research studies the robustness of multimodal models by making imperceptible adversarial changes to the individual input modalities using unimodal perturbations~\cite{li2020closer, chen2020counterfactual}. However, while adversarial perturbations to images are often deemed as imperceptible to humans, the adversarial perturbations in text often compromise the semantic meaning and its category to notable extents 
~\cite{wang2021adversarial}. In the context of multimodal learning, the problem of introducing textual perturbations that lead to semantically poor changes has been tackled by developing careful automated approaches -- for instance, by synthesizing counterfactual samples using language models~\cite{chen2020counterfactual}, or by conducting human-in-the-loop curation of adversarial examples~\cite{sheng2021human, li2021adversarial}. However, these studies only focus on VQA~\cite{antol2015vqa}. Additionally, as \citeauthor{gilmer2018motivating} (\citeyear{gilmer2018motivating}) argue, the imperceptibility criterion does not constrain the plausible action space in human-facing applications. For instance, it has been shown that the human-provided description of an image can vary notably with the personality, age, and location of the writer in terms of its length, emotion, and vocabulary; all the while preserving the cross-modal semantic interaction~\cite{shuster2019engaging, chunseong2017attend, denton2015user}.  Consequently, in this work, we focus on the robustness of multimodal classifiers to \textit{plausible} variations, specifically cross-modal dilutions. 

\begin{figure*}
    \centering
    \includegraphics[width = 1.0\textwidth]{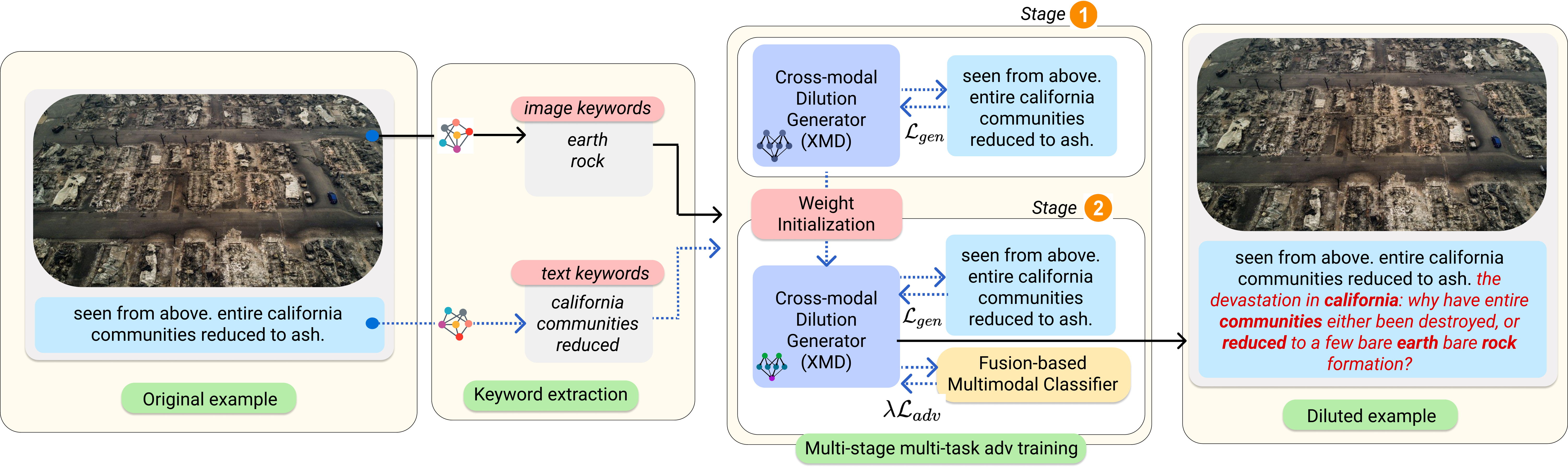}
    \caption{{\textbf{Overview of our proposed method.} We propose XMD --- Cross-Modal Dilution Generator. Our approach extracts keywords from the image and text of a multimodal example and generates dilution text that causes incorrect classifications by the multimodal classifier when appended to the original text. The generation model is trained in a multi-stage multi-task setup, where the adversarial loss component (stage 2) encourages the generation of dilution words that cause incorrect categorization. The blue dashed lines depict the training pathway.}  }
    \label{fig:method_figure}
\end{figure*}

\vspace{-1mm}
\noindent\textbf{Adversarial Perturbations}: \label{sec:related-adversarial-perturbations}
Our investigation concerns adding related text in a multimodal example to the existing textual information. 
Several methods have been proposed to introduce \textit{imperceptible} and \textit{adversarial} perturbations in text~\cite{li2021contextualized, li2020bert, garg2020bae}, focusing on word-level or phrase-level automated insertions, replacements, and merging.
Moving beyond the imperceptibility constraint, to estimate robustness to perceptible but plausible changes in text, recent research has investigated the robustness of NLP models to \textit{rule-based} distractions that are added to the original text~\cite{naik2018stress, ribeiro2020beyond}. As the constraints that govern textual dilutions in a multimodal setting are different, we propose a model to generate cross-modal (image $\rightarrow$ text) dilutions that maintain semantic and topical coherence with the existing image and text, while also demonstrating adversarial properties with respect to the multimodal classifiers. This provides us with a realistic estimate of the robustness of multimodal classifiers. 

\section{Cross-Modal Dilutions} \label{sec:distractions}
\vspace{-2mm}
Related work on language-only models~\cite{naik2018stress, ribeiro2020beyond} inspires us to study the robustness of deep multimodal classifiers to dilution. In the context of multimodal learning, dilutions can be introduced by adding information from the associated image to the original text. Since multimodal fusion models are expected to consider the information in images and text jointly, they should, in principle, be robust to the expression of additional information regarding the image in the form of text. This is, however, challenging to study because a plausible dilution should have semantic similarity with both the image and the original text. While a rule-based dilution like ``and true is true" (investigated by ~\citeauthor{naik2018stress} (\citeyear{naik2018stress})) are plausible for specific language-only tasks like Natural Language Inference~\cite{bowman2015large}, they do not cover the action space of plausible cross-modal dilutions for multimodal content.
Therefore, we develop an approach to generate dilutions that are semantically aligned with original text and image.

Our proposed approach follows the following framework to generate dilutions; see Figure \ref{fig:method_figure}.\\
\textit{\textbf{(i)}} Extract keywords from image and text based on their prominence in their respective modalities. \\
\textit{\textbf{(ii)}} Train a language model to fill words around the extracted keywords from the original text to generate dilutions~\cite{zhang2020pointer}. The generation model is trained using a multi-stage multi-task approach. The first stage fine-tunes the model to generate in-domain text using textual keywords in a self-supervised manner. The second stage involves training the model on an objective that combines generation loss with adversarial loss.\\
\textit{\textbf{(iii)}} The trained model is then used to generate text based on the keywords combined from both text and image modalities. The generated text is then appended to the original text as dilution. 

\subsection{Method for Generating Dilutions}
\label{sec:method_dilution}
\noindent \textbf{Multimodal classifier}: We design a fusion-based multimodal classification model ($\mathcal{M}_{mm}$) following  widely adopted architectures in both academic research and industrial applications~\cite{agarwal2020crisis, Dataminr}.
$\mathcal{M}_{mm}$ takes the concatenation of modality-specific representations as input and makes a joint classification. To model individual modalities, we first train an image-only classifier $\mathcal{M}_{image}$ and a text-only classifier $\mathcal{M}_{text}$ for the same classification task.
We then concatenate the output of the penultimate layers of the modality-specific models to feed them into a fully-connected network that is trained to fuse the modality-specific representations to perform joint classification based on the multimodal input.  

\noindent\textbf{Keyword extraction}: Our dilution generation approach is centered around keywords in the original image and text as that will ensure semantic relatedness of the dilution text with both the associated modalities. We use Yet Another Keyword Extractor (YAKE)~\cite{campos2018yake} to extract the most important keywords from the textual description for each example. For extracting keywords from the image, we consider the top $150$ objects in the Visual Genome dataset~\cite{krishna2017visual} and identify these objects in our dataset using a pre-trained image to Scene Graph generator~\cite{tang2020unbiased}. We further filtered the list of all identified objects by only considering objects with a bounding box that occupies at least $10\%$ of the total image area to ensure prominence in the image. These objects are considered the keywords of the image. We denote the keywords from text and image as $K_{text}$ and $K_{image}$, respectively. 

\noindent\textbf{Constrained text generation}: Once we have the keywords from text and image for each of the examples, the goal is to generate dilution around these keywords. For this, we extend the constrained text generation approach proposed by ~\citeauthor{zhang2020pointer} (\citeyear{zhang2020pointer}). We fine-tune a BERT language model to progressively predict [MASK] tokens around the initial set of keywords until only a special token (i.e., no-insertion token [NOI]) is predicted at all places to indicate no further insertions. 
We consider the original descriptions of the training examples in our target dataset and fine-tune the pre-trained model to reconstruct the original examples using keywords in the text, i.e., $K_{text}$. We adopt the same generation objective as ~\citeauthor{zhang2020pointer} (\citeyear{zhang2020pointer}) and denote it as $\mathcal{L}_{gen}$. The fine-tuned model can generate domain-specific text using the supplied keywords during inference. 

\noindent\textbf{Adversarial training}: While the above fine-tuning enables constrained generation of target-domain text based on the supplied keywords, we need to ensure that the generated dilutions also cause incorrect classifications by the trained multimodal classifier $\mathcal{M}_{mm}$. Explicitly designing the generation process to exhibit adversarial nature provides an estimate of the possible drop in performance in the presence of cross-modal dilutions. To this end, we consider the POINTER model after domain-specific fine-tuning and fine-tune it further using a combined loss function. The combined loss function takes into account not only the original generation loss but also a weighted component of the adversarial loss $\mathcal{L}_{adv}$. 
More formally,
\begin{equation}
    \mathcal{L}_{combined} = \mathcal{L}_{gen} + \lambda \mathcal{L}_{adv}
    \label{eq:loss_formulation}
\end{equation}
where $\lambda$ controls the contribution of the adversarial loss towards the generation process. The incorporation of $\mathcal{L}_{adv}$ encourages the generation model to fill the [MASK] tokens with words that would cause incorrect classifications by the multimodal classifier $\mathcal{M}_{mm}$. More formally, $\mathcal{L}_{adv}$ is computed for each training example as:
\[  
     \mathcal{L}_{adv} = - (y \log(\widehat{y}) + (1 - y) \log(1 - \widehat{y}))
\]
where $y=1$ when the predicted class by $\mathcal{M}_{mm}$ is different from the ground-truth class and $y=0$ when the predicted and ground-truth labels are the same. 
The probability of incorrect classification, i.e., $\hat{y}$, is obtained by adding the class probabilities of incorrect classes~\cite{le2020malcom, he2021petgen}. 
The training of the generation model is done in a multi-stage manner --- in the first stage, the model is fine-tuned to generate related in-domain text from keywords using $\mathcal{L}_{gen}$ and then in the second stage, it is trained in a multi-task fashion using a weighted combination of $\mathcal{L}_{gen}$ and  $\mathcal{L}_{adv}$. This ensures that the model maintains the quality and coherence in the generated text while learning adversarial behavior.

\noindent\textbf{Inference-time dilution generation}: We use the constrained text generation model described above to generate text based on combined keywords from both text and image (i.e., $K_{text} \oplus K_{image}$ --- where $\oplus$ denotes the concatenation of keywords). These generated textual dilutions are added to the original text to obtain examples with cross-modal dilutions. Our evaluation aims to assess the impact of these cross-modal dilutions on the performance of the trained multimodal classifier $M_{mm}$ along with various attributes of the generated text. 

\section{Multimodal Datasets}
\label{sec:datasets}
We conduct experiments on two user-generated datasets that have real-world societal applications. 

\noindent\textbf{Crisis Humanitarianism Dataset}: During crises, affected parties often use social media to communicate with humanitarian organizations that process the available information to provide timely and effective interventions. To aid development of related computational methods, ~\citeauthor{alam2018crisismmd} (\citeyear{alam2018crisismmd}) curated the CrisisMMD dataset.
This multimodal dataset comprises $7,216$ Twitter posts in English (images + text) that are categorized into $5$ humanitarian categories.\footnote{{\small{infrastructure and utility damage: 10\%}, {rescue volunteering or donation effort: 14\%}, {affected individuals: 1\%}, {other relevant information: 22\%}, \& {not humanitarian: 53\%}}}
We formulate the task of humanitarian information detection as a multi-class classification problem, and use the standard training ($n = 5263$), evaluation ($n = 998$), and test ($n = 955$) sets in our experiments. 

\noindent\textbf{Sentiment Detection Dataset}: User-generated content has been frequently used to infer sentiments of individuals for various applications, including detection of mental health indicators~\cite{de2013social}. We collect the dataset introduced by ~\citeauthor{duong2017multimodal} (\citeyear{duong2017multimodal}) for the task of sentiment detection. The dataset comprises multimodal posts (in English) from Reddit that are categorized into 4 classes.\footnote{\small{{creepy: 22\%}, {rage: 19\%}, {gore: 25\%}, \& {happy: 34\%}}} We crawled the images from Reddit URLs provided by the authors and split the dataset in a 80:10:10 ratio to obtain the train ($n = 2568$), validation ($n = 321$), and test ($n = 318$) sets. 

\section{Experiments}
We first discuss the training of our proposed cross-modal dilutions (XMD) generator model. Then, we discuss multiple baselines that dilute the original text using various rule- and model-based approaches. Finally, we evaluate XMD and compare its performance with the baselines.

\subsection{Training Details}
\noindent\textbf{Multimodal Classifier} (${\mathcal{M}_{mm}}$): ${\mathcal{M}_{mm}}$ is a fusion of text-only and image-only classifiers.
For text-only classifier ${\mathcal{M}_{text}}$, we fine-tune and evaluate a BERT~\cite{devlin2018bert} model on the target dataset. Similarly, we fine-tune a VGG-16 model~\cite{simonyanVeryDeepConvolutional2015a} pre-trained on ImageNet~\cite{imagenet2009Deng} to train an image-only classifier ${\mathcal{M}_{image}}$. We refer the reader to Appendix \ref{sec:app_text_classifier} and \ref{sec:app_image_classifier} for details and evaluation of the modality-specific classifiers.
 We feed the concatenation of fine-tuned text and image representations to the multimodal classifier, which is essentially a series of fully-connected layers with ReLU activation~\cite{agarap2018deep}. The architecture of the multimodal classifier comprises an input layer ($1024$ neurons), $3$ hidden layers ($512$, $128$, $32$ neurons), and an output layer (neurons $=$ number of classes in the dataset). We use Adam optimizer~\cite{kingma2014adam} with a learning rate initialized at $10^{-4}$ and adopt early stopping based on the validation set loss to avoid overfitting. 
 
\noindent\textbf{Cross-modal dilutions generator} (XMD): 
For keyword extraction from YAKE, we set the maximum n-gram size to $1$, the de-duplication threshold to $0.9$ with `seqm' function, and the window size to $1$. The rest of the hyper-parameters were set to their default values used in previous studies~\cite{zhang2020pointer, tang2020unbiased}.
We fine-tune the POINTER model pre-trained on Wikipedia text~\cite{zhang2020pointer} using default hyper-parameters for $5$ epochs. During this stage of the training, the objective is $\mathcal{L}_{gen}$ and the model learns to generate text from keywords that aligns with the target domain. Following this, we further train the generation model for another $1$ epoch using the combined objective in Equation \ref{eq:loss_formulation}, while setting $\lambda = 0.01$ (based on results on the validation set). This adversarial adaptation of the model encourages generations that could cause misclassifications by the trained $\mathcal{M}_{mm}$. Finally, the keywords from images and text (i.e.,  $K_{text} \oplus K_{image}$) are passed as input to XMD to generate dilutions for the examples in the test set.

\subsection{Baselines}
\noindent\textbf{Rule-based dilutions}:\\
\textit{(i) Random URL}: As proposed by ~\citeauthor{ribeiro2020beyond} (\citeyear{ribeiro2020beyond}), we append a randomly generated twitter URL (e.g., https:t.co/gXvDrs) to the original text. \\
\textit{(ii) Relevant keywords}: We experiment with adding extracted keywords from the image, text, and both together to the original text.\\
\textit{(iii) Most similar image's description}: We add the textual description of the most similar image (computed using cosine similarity between fine-tuned VGG-16 embeddings of images in the test set) to the original text. This mimics scenarios where the user dilutes the original text by adding the description of a highly similar image; see Appendix \ref{sec:app_mismatch_baseline}. 

\vspace{0.05in}
\noindent\textbf{Model-based dilutions}:\\
\textit{(i) GPT}: We use the original text as the prompt for a GPT-2~\cite{radford2019language, wolf2020transformers} model and add the generated text to it for dilution. \\
\textit{(ii) GPT Fine-tuned}: We first fine-tune a GPT-2 model using the text in the training set of the dataset (using default hyper-parameters) for domain adaptation, and then use the original text as the prompt to obtain the dilution text.\\
\textit{(iii) Image Captioning}: We use two trained image captioning models (SCST~\cite{rennie2017self} \& XLAN~\cite{pan2020x}) to generate the captions for the images in the test set. We append the generated captions to the original text for dilution.

\subsection{Evaluation metrics}
Our evaluation is focused on assessing two aspects of the dilutions: \textit{(a)} are the dilutions effective in deteriorating the classification performance of the multimodal classifier?, and \textit{(b)} are the added dilutions relevant to the original text + image, and maintain topical coherence with the existing text? To this end, we compute standard classification metrics for the former evaluation and compute embedding-based similarity measures for the latter. \textbf{Sim}$_{text}$ denotes the similarity between the original text and the generated dilution and is computed using the cosine similarity between the embeddings from the fine-tuned BERT classifier. Similarly,  \textbf{Sim}$_{img}$ denotes the similarity between the generated dilution and the image and is computed using cosine similarity between CLIP embeddings~\cite{radford2021learning}. For topical coherence, we compute the KL Divergence (\textbf{KL Div}) between the topic distributions of the original text and the generated text. For details regarding the training of the topic model, please see Appendix \ref{sec:app_topic_model}. Additionally, we quantify the correspondence similarity between image and final text (i.e., original + dilution) using a learned metric \textbf{Sim}$_{corr}$ that quantifies the correspondence between \textit{diluted descriptions and original images} based on the correspondence between \textit{original text and images}; see Appendix \ref{sec:app_corr_similarity} for further details. 
Furthermore, we compute Self-BLEU~\cite{zhu2018texygen} scores for the sentences in the generated dilution to quantify diversity, wherever applicable. For all model-based baselines and XMD, we report the average values over $5$ runs with different random seeds. 

\begin{table*}[t!]
    \centering
    \resizebox{0.90\linewidth}{!}{%
    \begin{tabular}{
    l  c c c c  c c c  c  c}
    \toprule
       & \multicolumn{4}{c}{\sc Classification Performance $\downarrow$} & \multicolumn{3}{c}{\sc Relevance $\uparrow$} & \textsc{Diversity $\downarrow$} & \textsc{Topical Diff. $\downarrow$}\\
      \cmidrule(lr){2-5}
      \cmidrule(lr){6-8}
      \cmidrule(lr){9-9}
      \cmidrule(lr){10-10}
        & $\mathbf{F_1}$ & $\mathbf{Prec.}$ & $\mathbf{Recall}$ & $\mathbf{Acc.}$ & 
        $\mathbf{Sim_{\bf \it text}}$ & $\mathbf{Sim_{\bf \it img}}$ & $\mathbf{Sim_{\bf \it corr}}$ & \textbf{Self-BLEU} & \textbf{KL Div.}\\
        \midrule
        \textbf{Original} & 0.734 & 0.742 & 0.725 & 0.828 & -- & 0.292 & 0.999 & 0.048 & --\\  
        \midrule
        \textbf{Rule-based} & & & & & & & & &\\
        \hspace{3mm} Random URL & 0.705 & 0.747 & 0.672 & 0.817 & 0.467 & -- & 0.967 & -- & --\\
        \hspace{3mm} Image KW & 0.733 & 0.735 & 0.757 & 0.822 & 0.498 & 0.194 & \textbf{0.989} & -- & 4.383\\
        \hspace{3mm} Text KW & 0.736 & 0.744 & 0.736 & 0.823 & 0.703 & \textbf{0.233} & 0.991 & -- & \textbf{2.022}\\
        \hspace{3mm} Text + Image KW & 0.706 & 0.716 & 0.702 & 0.824 & 0.656 & 0.232 & 0.988 & -- & 4.618\\
        \hspace{3mm} Similar image's desc & 0.642 & 0.624 & 0.677 & 0.751 & 0.597 & 0.204 & 0.962 & 0.049 & 10.104\\  
        \midrule
        \textbf{Model-based} & & & & & & & & &\\
        \hspace{3mm} GPT & 0.684 & 0.693 & 0.677 & 0.783 & 0.562 & 0.221 & 0.971 & 0.081 & 9.251\\
        \hspace{3mm} GPT-FT & 0.628 & 0.616 & 0.629 & 0.754 & 0.614 & 0.216 & 0.981 & 0.063 & 8.182\\
        \hspace{3mm} SCST Captions & 0.666 & 0.696 & 0.663 & 0.774 & 0.502 & 0.200 & 0.979 & -- & 11.443\\
        \hspace{3mm} XLAN Captions & 0.673 & 0.703 & 0.677 & 0.782 & 0.534 & 0.218 & 0.980 & -- & 10.733\\
        \midrule
        \textbf{XMD (Ours)}  & \textbf{0.564} &\textbf{ 0.571} & \textbf{0.552} & \textbf{0.718} & \textbf{0.715} & 0.232 & 0.985 & \textbf{0.035} & 6.113\\
    \bottomrule
    \end{tabular}%
    }
    \caption{{\textbf{Results on the multimodal Crisis Humanitarianism dataset.} We evaluate dilution methods based on classification performance (lower values denote greater adversarial effectiveness of dilutions), relevance (higher similarity scores denote more relevance), diversity (lower Self-BLEU score denote more diverse sentences in generation), and topical differences (lower KL Divergence denotes better topical coherence). Our proposed method is compared against rule-based and model-based baselines.}}
    \label{tab:results-crisis}
\end{table*}

\begin{table}[!h]
    \centering
    \resizebox{1.0\linewidth}{!}{%
    \begin{tabular}{
    l c  c c c  c}
    \toprule
       & \multicolumn{1}{c}{\sc $F_1$ $\downarrow$} & \multicolumn{3}{c}{\sc Relevance $\uparrow$} & \textsc{Topical Diff. $\downarrow$}\\
      \cmidrule(lr){3-5}
      \cmidrule(lr){6-6}
        & & 
        $\mathbf{Sim_{\bf \it text}}$ & $\mathbf{Sim_{\bf \it img}}$ & $\mathbf{Sim_{\bf \it corr}}$ & \textbf{KL Div.}\\
        \midrule
        \textbf{Original} & 0.793 & -- & 0.314 & 0.999 & --\\  
        \midrule
        \textbf{Rule-based} & & & & &\\
        \hspace{3mm} Random URL & 0.773  & 0.538 & -- & 0.960 &  --\\
        \hspace{3mm} Image KW & 0.783 & 0.559 & 0.204 & 0.983 & 5.163\\
        \hspace{3mm} Text KW & 0.792 & 0.834 & 0.231 & \textbf{0.992} & \textbf{3.114}\\
        \hspace{3mm} Text + Image KW & 0.774 & 0.689 & 0.231 & 0.988 & 5.877\\
        \hspace{3mm} Similar image's desc & 0.665 & 0.611 & 0.268 & 0.963 & 11.980\\  
        \midrule
        \textbf{Model-based} & & & & &\\
        \hspace{3mm} GPT & 0.691 & 0.620 & 0.274 & 0.978 & 11.638\\
        \hspace{3mm} GPT-FT & 0.652 & 0.642 & 0.261 & 0.981 & 10.091\\
        \hspace{3mm} SCST Captions  & 0.671  & 0.553 & 0.251 & 0.971 & 13.427\\
        \hspace{3mm} XLAN Captions  & 0.651 & 0.588 & 0.261 & 0.979 &  14.612\\
        \midrule
        \textbf{XMD (Ours)}  & \textbf{0.614} & \textbf{0.795} & \textbf{0.298} & 0.984 & 9.137\\
        
    \bottomrule
    \end{tabular}%
    }
    \caption{{\textbf{Results on the multimodal Sentiment Detection.} Similar trends on a different dataset reinforce the adversarial effectiveness of XMD while generating relevant and coherent dilutions. Complete results are presented in Appendix \ref{sec:app_sentiment_results}.}}
    \label{tab:results-sentiment}
\end{table}

\section{Main Results}\label{sec:results}
Our results (Tables \ref{tab:results-crisis} \& \ref{tab:results-sentiment}) show the following:\\
$\bullet$ Rule-based dilutions do not demonstrate adversarial effectiveness with the exception of using most similar image's description as dilution, which however, shows poor relevance and topical coherence.\\
$\bullet$ Model-based baselines show adversarial effectiveness but lack in relevance and coherence.\\
$\bullet$ XMD demonstrates the best adversarial effectiveness while generating more relevant and topically coherent dilutions with respect to all the baselines.\\
$\bullet$ Our results generalize over both the datasets under considerations --- see Table \ref{tab:results-crisis} for Crisis Humanitarianism and Table \ref{tab:results-sentiment} for Sentiment Detection.
We elaborate on these results next.

\vspace{0.05in}\noindent\textbf{Effect of rule-based baseline dilutions}: We start by noting that the insertion of random URL, keywords from image, text, and both together, are ineffective in decreasing the classification performance of multimodal classifiers considerably.  
However, inserting the most similar image's description to the original text substantially lowers the classification performance, from $F_1$ score of 0.734 to 0.642 (12.3\% drop) for Crisis Humanitarianism dataset and from 0.793 to 0.665 (16.4\% drop) for the Sentiment Detection dataset. This indicates that adding text corresponding to a similar image in the dataset is a reasonably effective dilution strategy. However, since the most similar image in the dataset could correspond to a different class, using its description as dilution frequently leads to less relevance and low topical coherence, as indicated by low values of \textbf{Sim}$_{text}$, \textbf{Sim}$_{img}$, and \textbf{KL Div}.

\noindent\textbf{Effect of model-based baseline dilutions}: Model-based baseline dilution strategies are generally more effective than rule-based dilution strategies in lowering the classification performance of the multimodal models. The drop in $F_1$ scores ranges from 9.6\% (0.734 $\rightarrow$ 0.684) using GPT to 15.1\% (0.734 $\rightarrow$ 0.628) using GPT-FT for the Crisis Humanitarianism dataset. Similar trends are observed on the Sentiment Detection dataset.
Since GPT-FT is fine-tuned on in-domain text, the inserted text demonstrates a higher relevance with the original text when compared to GPT alone. Similarly, consistently across the two datasets, the correspondence similarity and the topical similarity scores for GPT-FT based dilutions are better than those of GPT. While the caption generation-based dilution strategies are also effective, they show lower relevance with existing text and a higher topical difference due to domain mismatch. The generated captions are generic and do not cater to the domains of crises and sentiment. Given the performance of all the model-based baselines across all the metrics, we consider GPT-FT to be the most competitive baseline. Overall, these results demonstrate that model-based baseline dilutions, whether text-only (GPT and GPT-FT) or cross-modal (using SCST and XLAN caption generation models), severely affect the performance of multimodal classifiers but lack in terms of relevance and coherence.

\vspace{0.05in}
\noindent\textbf{Effect of proposed cross-modal dilutions}: The cross-modal dilutions added using our approach lead to a drop in $F_1$ scores from 0.734 to 0.564 (23.3\%) and from 0.793 to 0.614 (22.5\%) for the Crisis Humanitarianism and Sentiment datasets, respectively. This is by far the most effective dilution strategy that also demonstrates high relevance with the original text and image, high correspondence similarity, and low topical difference. The observed trends are consistent across both datasets. The superior performance of XMD across all metrics can be attributed to several design choices. First, XMD is designed to exploit the model vulnerabilities by encouraging misclassification via an adversarial loss component in the training objective. Second, while dilution using GPT-FT only considers the original text as context, XMD generates text based on the keywords from both the original text and the image. This results in relatively higher relevance to the original text and the image.
Finally, even though both GPT-FT and XMD are trained to generate in-domain text via task-specific fine-tuning, XMD exceeds in terms of topical similarity between inserted and original text ({KL Div}: 6.113 versus 8.182 for Crisis Humanitarianism dataset).

It is worth mentioning that our proposed method (XMD) also generates text with the highest diversity across generated sentences compared to all the baselines. This is demonstrated by lowest Self-BLEU scores in Tables \ref{tab:results-crisis} and \ref{tab:results-sentiment}. However, since the values for all the methods are consistently small, all the dilutions can be considered sufficiently diverse. 

To summarize, we observe that deep multimodal classifiers are not overly sensitive to minor content dilutions like the insertion of random URLs or keywords from the original content. However, adding dilutions based on text-alone (GPT, GPT-FT) or cross-modal (Captions, XMD) causes a notable drop in the classification performance of multimodal models. To this end, our proposed XMD generates the most effective dilutions in terms of the observed drop in classification performance while maintaining relevance with the original image and text and topical coherence. 

\section{Analysis of Cross-Modal Dilutions}
\vspace{-1mm}
Next, we further analyze the dilutions generated by our proposed method (XMD). We focus on the Crisis Humanitarianism dataset for our analyses.
In addition to the analyses presented here, we investigate the effect of the length of dilutions (i.e., number of inserted words) on classification performance and observe no notable difference in observed trends with similar dilution lengths; see Appendix \ref{sec:app_length_analysis}. In Appendix \ref{sec:app_lambda_variation}, we analyze the sensitivity of quantified metrics with respect to variations in $\lambda$. Finally, we conduct a human evaluation to assess how realistic the \textit{diluted} multimodal examples are when compared against \textit{real} multimodal examples.  

\begin{figure*}
    \centering
    \includegraphics[width=1.0\textwidth]{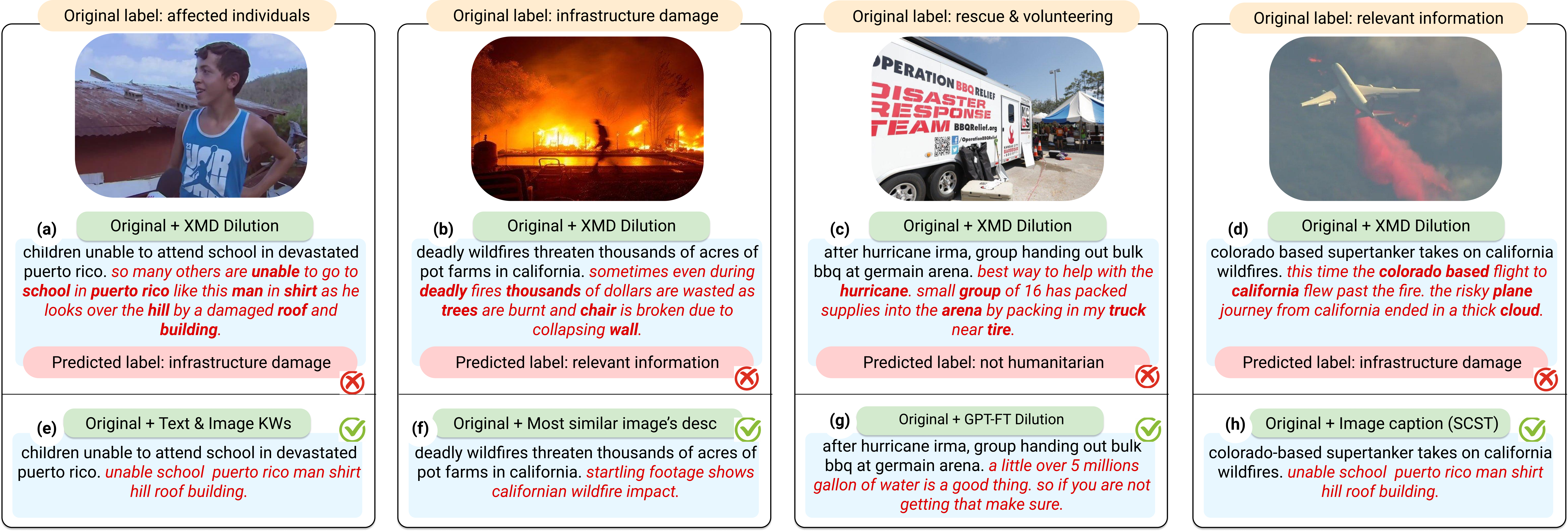}\vspace{-1mm}
    \caption{{\textbf{Qualitative examples of XMD Dilutions and baselines.} \textit{(a-d)}: Examples that are misclassified by the multimodal classifier \textit{after} adding dilutions generated by XMD; the original examples were classified correctly. \textit{(e-h)}: For each example, we also show what a baseline method would have added as a dilution which did not lead to incorrect classification. The original text is shown in black and the inserted dilutions are shown in red; extracted keywords are in bold.}}
    \label{fig:qualitative_examples}
\end{figure*}

\vspace{0.05in}\noindent\textbf{Subjective Assessment of Dilutions}: Figure \ref{fig:qualitative_examples} shows examples of the dilutions generated by XMD from the Crisis Humanitarianism dataset along with dilutions obtained from the baselines. 

To further assess the quality of generated dilutions, we conducted a survey on Amazon Mechanical Turk (AMT). We instructed the annotators to compare two multimodal posts --- one containing dilutions from XMD and the other containing dilutions from GPT-FT for the same multimodal example. The posts were randomly ordered to mitigate position bias. Annotators were asked to respond on a 5-point Likert scale (1: strongly disagree, 5: strongly agree) to the following question: \textit{Based on the quality of the text and its relevance with the image, is the post on the right more likely to be an actual social media post than the post on the left?} We obtained 5 annotations for each of the 200 examples that were randomly sampled from the test set of the Crisis Humanitarianism dataset. Overall, the results showed that annotators consider the dilutions generated by XMD to be more realistic than GPT-FT. The percentage of examples for which the majority of annotators preferred XMD dilutions over GPT-FT dilutions were: $32.1\%$ (strongly) and $46.4\%$ (moderately). For $12.2\%$ examples, the majority of annotators were neutral and preferred GPT-FT dilutions over XMD dilutions for $9.3\%$ examples. See Appendix \ref{sec:app_human_evaluation} for more details about human evaluation, recruitment, and compensation. 

\vspace{0.03in}
\noindent\textbf{Ablations}: 
We aim to understand the role of two key components in XMD -- the incorporation of the adversarial loss component in the training objective and the inclusion of textual keywords. For the Crisis Humanitarianism dataset, Table \ref{tab:results-crisis-ablation} shows that \textit{(a)} XMD without the adversarial loss component and without infusing keywords from the original text (i.e., XMD (Plain)) lacks in generating relevant and topically coherent dilutions. On adding adversarial loss component to the objective (i.e., XMD (Adv)), the classification performance decreases further with little effect on relevance and coherence. Keeping the adversarial loss while infusing keywords from the original text (i.e., XMD (Full)) leads to the largest drop in classification performance while improving relevance (with both text and image) as well as topical coherence. Ablations on the Sentiment Detection task show the same trends; see Appendix \ref{sec:app_ablation_sentiment}. 

\begin{table}[!t]
    \centering
    \resizebox{\columnwidth}{!}{%
    \begin{tabular}{
    l  c c c c  c c  c}
    \toprule
       & \multicolumn{4}{c}{\sc Classification Performance $\downarrow$} & \multicolumn{2}{c}{\sc Relevance $\uparrow$} &  \textsc{Top. Diff. $\downarrow$}\\
      \cmidrule(lr){2-5}
      \cmidrule(lr){6-7}
      \cmidrule(lr){8-8}
        & $\mathbf{F_1}$ & $\mathbf{Prec.}$ & $\mathbf{Recall}$ & $\mathbf{Acc.}$ & 
        $\mathbf{Sim_{\bf \it text}}$ & $\mathbf{Sim_{\bf \it img}}$  & \textbf{KL Div.}\\
        \midrule
        \textbf{XMD (Plain)}  & 0.643 & 0.656 & 0.651 & 0.770 & 0.493 & 0.195 &  9.246\\
        \textbf{XMD (Adv)}  & 0.624 & 0.632 & 0.621 & 0.739 & 0.483 & 0.193 & 9.275\\
        \textbf{XMD (Full)}  & 0.564 & 0.571 & 0.552 & 0.718 & 0.715 & 0.232 & 6.113\\
        
    \bottomrule
    \end{tabular}}
    \caption{{\textbf{Ablation Results for Crisis Humanitarianism Dataset.} \textbf{Plain} is trained with $\mathcal{L}_{gen}$ alone and only uses image keywords. \textbf{Adv} is trained using $\mathcal{L}_{gen} + \lambda \mathcal{L}_{adv}$. \textbf{Full} includes adversarial training + text \& image keywords.}}
    \vspace{-1mm}
    \label{tab:results-crisis-ablation}
\end{table}

\vspace{0.05in}\noindent\textbf{Are cross-modal content dilutions realistic?}
We now focus on assessing how \textit{realistic} the diluted examples are when compared to the real social media examples. To this end, we conduct an AMT survey that requires users to compare multimodal examples with inserted dilutions against different but original multimodal examples. To prime the annotators, we first show them 5 unmodified multimodal examples and subsequently ask them to analyze a list of randomly-ordered multimodal examples, half diluted and the other half original unmodified examples. We use the dilutions generated by XMD. For each example in the list, the annotators are asked to respond to the following question on a 5-point Likert scale: \textit{Do you think this post (text and image) could be a real post from social media website?} We select a subset of 100 examples from the test set of the Crisis Humanitarianism dataset and obtain 3 annotations for each example. 
The average Likert score for original examples is 3.61 ($\pm$ 0.53), whereas that for diluted examples is 3.38 ($\pm$ 0.39). The inter-rater agreement indicated strong reliability of annotations (Krippendorf's $\alpha = 0.83$). An independent two-sided t-test (assuming unequal variances) resulted in a $p$-value of $0.24$, indicating no evidence that the average Likert scores of the original and diluted examples are from different distributions. These results show that the annotators asses the diluted and original examples to be similar, reinforcing the realistic nature of dilutions. 

\section{Conclusion and Future Work} \label{sec:analysis}
In sum, our work is the first investigation of the robustness of multimodal classifiers to cross-modal dilutions. We establish the plausibility of such dilutions via human evaluations and develop a model to emulate adversarial scenarios reliably. We find that multimodal classifiers that fuse the state-of-the-art modality-specific representations are not robust to cross-modal dilutions generated by XMD.

Deep classifiers are increasingly being used for crucial applications that involve the joint understanding of user-generated multimodal data. Our broader goal in this work is to analyze and advocate for the robustness of multimodal models with societal applications, while focusing on the most representative fusion-based multimodal classification technique. 
In the future, we intend to leverage the knowledge of vulnerabilities identified in the current work to develop more robust multimodal models. 
We encourage interested researchers to investigate other cross-modal variations pertinent to multimodal data and assess the robustness of multimodal learning approaches to these variations.

\section{Limitations and Broader Perspective}

It is important to be clear about the limitations of this work. Our approach hinges on extracting informative keywords from both the image and the text to ensure the relevancy of the generated dilutions. In scenarios where the extracted keywords from images are generic (like celebrity faces for multimodal fake news detection) or the contextual relationship between image and text modalities is not straightforward (like multimodal hate speech), the proposed method does not generate semantically meaningful dilutions.  
We discuss the limitations in greater detail in Appendix \ref{sec:app_limitations}.  

This work emphasizes the possibility that the lack of robustness of multimodal classification models can cause societal harm, such as delaying humanitarian interventions during crisis events. 
As such, the trained adversarial dilution generation models could be put to malicious use. We strongly condemn the misuse of this research. We release the code to aid reproducibility and promote future research on this topic. We believe that this research will encourage the community to investigate the robustness of multimodal classifiers and minimize real-world harm, leading to long-term benefits. 

\vspace{0.01in}\noindent\textit{Bias of pre-trained models}: It is known that pre-trained models used in our study demonstrate many biases~\cite{bender2019rule, hendricks2018women, garimella2021he}. This is often reflected in the kind of keywords that are identified in images and the resulting generated text (e.g., stereotypical gender associations). We acknowledge that the current state of deep learning research is limiting, and the consequential shortcomings are reflected in our work to some extent. 

\vspace{0.01in}\noindent\textit{Annotations, IRB approval, and datasets}: The annotators for this study were recruited via AMT. We specifically recruited `Master' annotators located in the United States; and paid them at an hourly rate of 10 USD for their annotations. The human evaluation experiments were approved by the Institutional Review Board at Georgia Tech. The datasets used in this study are publicly available and were curated by previous research.

\vspace{-1mm}
\section{Acknowledgements}
\vspace{-1mm}
This research/material is based upon work supported in part by the Defense Advanced Research Projects Agency (DARPA) under Agreement No. HR00112290102 (subcontract No. PO70745), NSF ITE-2137724, NSF ITE-2230692, Microsoft AI for Health, IDEaS at Georgia Institute of Technology, and Adobe Inc. Any opinions, findings and conclusions or recommendations expressed in this material are those of the author(s) and do not necessarily reflect the position or policy of DARPA, DoD, NSF, and SRI International and no official endorsement should be inferred. We thank Shivaen Ramshetty for sharing insights from related experiments during the rebuttal phase, the CLAWS research group members for their inputs, and the anonymous reviewers for their constructive feedback.
\balance
\bibliography{custom}
\bibliographystyle{acl_natbib}

\appendix

\section{Appendix}
\label{sec:appendix-additional-results}

\subsection{Text-only Classifier Training}
\label{sec:app_text_classifier}

Before training, we pre-process the text in multimodal examples to remove URLs, emoticons, platform-specific tokens (like `RT' for indicating retweets on Twitter), and symbols like @ and \#. We also expanded negatives like \textit{can't} and \textit{won't} to `\textit{can not}' and `\textit{will not}'. To train the text classifier (${\mathcal{M}_{text}}$), we fine-tune a pre-trained language model,  DistilBERT~\cite{sanh2019distilbert, devlin2018bert}, on the two datasets discussed in Section \ref{sec:datasets} by using the respective training sets. To train the text classification models for each dataset, we use Adam optimizer~\cite{kingma2014adam} with a learning rate initialized at $10^{-4}$; hyper-parameters are set by observing the classification performance achieved on the respective validation set. We use early stopping~\cite{caruana2000overfitting} to stop training when the loss value on the validation set stops to improve for 5 consecutive epochs. The performance of the trained classifier on the test sets of Crisis Humanitarianism and Sentiment Detection datasets are presented in Table \ref{tab:modality_specific_results}.

\begin{table}[!b]
    \centering
    \resizebox{\columnwidth}{!}{%
    \begin{tabular}{
    l  c c c c }
    \toprule
      \mystar \textbf{ Crisis Humanitarianism} & \multicolumn{4}{c}{\sc Classification Performance}\\
      \cmidrule(lr){2-5}
        & $\mathbf{F_1}$ & $\mathbf{Prec.}$ & $\mathbf{Recall}$ & $\mathbf{Acc.}$ \\
        \midrule
        \textbf{Text-only Classifier}  & 0.713 & 0.725  & 0.703  & 0.801 \\
        \textbf{Image-only Classifier}  & 0.429  & 0.456  & 0.426  & 0.528\\
        \textbf{Multimodal Classifier}  & 0.734 & 0.742 & 0.725 & 0.828 \\
        
    \bottomrule
    \bottomrule
    \toprule
      \mystar \textbf{ Sentiment Detection} & \multicolumn{4}{c}{\sc Classification Performance}\\
      \cmidrule(lr){2-5}
        & $\mathbf{F_1}$ & $\mathbf{Prec.}$ & $\mathbf{Recall}$ & $\mathbf{Acc.}$ \\
        \midrule
        \textbf{Text-only Classifier}  & 0.732 & 0.739  & 0.733  & 0.742 \\
        \textbf{Image-only Classifier}  & 0.941  & 0.948  & 0.946  & 0.953\\
        \textbf{Multimodal Classifier}  & 0.793 & 0.797 & 0.798 & 0.802 \\
        
    \bottomrule
    \end{tabular}}
    \caption{{Performance of text-only and image-only classifiers on the Crisis Humanitarianism and Sentiment Detection tasks.}}
    \label{tab:modality_specific_results}
\end{table}

\subsection{Image-only classifier}
\label{sec:app_image_classifier}
We apply a standard image pre-processing pipeline so that images with different dimensions can fit the pre-trained VGG-16 model's input requirement. First, we resize the image so that its shorter dimension is $224$. We then crop the square region in the center and normalize the square image with the mean and standard deviation of the ImageNet images~\cite{imagenet2009Deng}.

To train the image-only classifier (${\mathcal{M}_{image}}$), we apply a fine-tuning approach to train the task-specific image classifiers. We first freeze the weights of VGG-16~\cite{simonyanVeryDeepConvolutional2015a}, pre-trained on ImageNet~\cite{imagenet2009Deng}, and  then swap the last layer from the original model to three fully connected hidden layers with dimensions \texttt{4096}, \texttt{256}, and \texttt{num-of-classes}.
Finally, we retrain these three layers to adapt the image distribution in each dataset.
We use Adam optimizer~\cite{kingma2014adam} with a learning rate of $10^{-4}$ for each dataset.
To avoid overfitting, we use early stopping to stop training when the loss value on the validation set stops to improve for 10 consecutive epochs. Table \ref{tab:modality_specific_results} shows the performance of image-only classifier.

\subsection{Keyword Extraction from YAKE}
\label{sec:app_kw_extractor}
For extracting keywords from the original text, we use YAKE~\cite{campos2018yake}. We set the following hyper-parameters: maximum N-gram size = 1; de-duplication threshold = 0.9; de-duplication algorithm: `seqm'; window size = 1, maximum number of keywords extracted from text = 5.

\subsection{Baseline: Most similar image's desc.}
\label{sec:app_mismatch_baseline}
We create this baseline to emulate the scenario where the user could have posted the multimodal example after diluting the original text by adding a highly similar image's description. We find the most similar image in the test set to an image of a given multimodal example and append its caption to the text in the given multimodal example. As mentioned in the main text, we use the cosine similarity between the VGG-16 embeddings obtained after task-specific for computing the similarity. Overall, most similar images were found to be highly similar, with an average highest similarity score of $0.767$ with a standard deviation of $0.067$. Nonetheless, as discussed in Section \ref{sec:results}, this naïve dilution strategy frequently leads to irrelevant and topically incoherent. 

\subsection{Evaluation: Correspondence similarity}
\label{sec:app_corr_similarity}
We explain the rationale behind adopting the correspondence similarity score (i.e., \textbf{Sim}$_{corr}$) as one of our evaluation metrics. For context, the cross-modal correspondence prediction task is a binary classification task that aims to classify two input modalities as corresponding or not. For instance, if an image and text that are parts of the same multimodal example are provided as input, the correct prediction is Label $1$, indicating true correspondence. Conversely, if the input text and image are from different multimodal examples, the correct prediction is Label $0$, indicating false correspondence. The correspondence prediction task has been widely adopted as a pre-training step for multimodal deep learning models~\cite{arandjelovic2017look, verma2019learning, feng2014cross}. In this work, we train correspondence prediction models using the fine-tuned image and text representations of the dataset-specific \textit{undiluted} training set, and then report \textbf{Sim}$_{corr}$ --- the average probability score for Label $1$ (i.e., true correspondence) on the \textit{diluted} dataset-specific test set examples. Effectively, the score indicates that given a model trained to predict correspondence between image and text from original unmodified training examples; the model is successful in establishing a correspondence between diluted text and images in the test set examples. 

To train the cross-modal correspondence prediction model, we create negative examples by randomly sampling $3$ mismatched descriptions from the training set for each image with the correct description. We then take the fine-tuned representation of the input image and text and pass them through a series of fully-connected layers of sizes (1024 (input), 512, 256, 128, 64, 32, and 2 (output)). As shown in Tables $\ref{tab:results-crisis}$ and \ref{tab:results-sentiment}, the correspondence prediction model provides a nearly-perfect \textbf{Sim} score (i.e., 0.999) on undiluted test sets. However, the scores for baselines and the proposed model differ based on the dilution strategy adopted. 

\begin{table*}[!t]
    \centering
    \resizebox{0.85\linewidth}{!}{%
    \begin{tabular}{
    l  c c c c  c c c  c  c}
    \toprule
       & \multicolumn{4}{c}{\sc Classification Performance $\downarrow$} & \multicolumn{3}{c}{\sc Relevance $\uparrow$} & \textsc{Diversity $\downarrow$} & \textsc{Topical Diff. $\downarrow$}\\
      \cmidrule(lr){2-5}
      \cmidrule(lr){6-8}
      \cmidrule(lr){9-9}
      \cmidrule(lr){10-10}
        & $\mathbf{F_1}$ & $\mathbf{Prec.}$ & $\mathbf{Recall}$ & $\mathbf{Acc.}$ & 
        $\mathbf{Sim_{\bf \it text}}$ & $\mathbf{Sim_{\bf \it img}}$ & $\mathbf{Sim_{\bf \it corr}}$ & \textbf{Self-BLEU} & \textbf{KL Div.}\\
        \midrule
        \textbf{Original} & 0.793 & 0.797 & 0.798 & 0.802 & -- & 0.314 & 0.999 & 0.053 & --\\  
        \midrule
        \textbf{Rule-based} & & & & & & & & &\\
        \hspace{3mm} Random URL & 0.773 & 0.777 & 0.772 & 0.784 & 0.538 & -- & 0.960 & -- & --\\
        \hspace{3mm} Image KW & 0.783 & 0.782 & 0.794 & 0.796 & 0.559 & 0.204 & 0.983 & --  & 5.163\\
        \hspace{3mm} Text KW & 0.792 & 0.791 & 0.798 & 0.801 & 0.834 & 0.231 & 0.992 & -- & 3.114\\
        \hspace{3mm} Text + Image KW & 0.774 & 0.771 & 0.768 & 0.785 & 0.689 & 0.231 & 0.988 &  -- & 5.877\\
        \hspace{3mm} Similar image's desc & 0.665 & 0.662  & 0.676 & 0.680 & 0.611 & 0.268 & 0.963 & 0.052 & 11.980\\  
        \midrule
        \textbf{Model-based} & & & & & & & & &\\
        \hspace{3mm} GPT & 0.691 & 0.695 & 0.683 & 0.697 & 0.620 & 0.274 & 0.978 & 0.086 & 11.638\\
        \hspace{3mm} GPT-FT & 0.652 & 0.664 & 0.661 & 0.668 & 0.642 & 0.261 & 0.981 & 0.074 & 10.091\\
        \hspace{3mm} SCST Captions  & 0.671 & 0.675 & 0.681 & 0.680 & 0.553 & 0.251 & 0.971 & -- & 13.427\\
        \hspace{3mm} XLAN Captions  & 0.651 & 0.663 & 0.656 & 0.665 & 0.588 & 0.261 & 0.979 & -- & 14.612\\
        \midrule
        \textbf{XMD (Ours)}  & 0.614 & 0.617 & 0.626 & 0.633 & 0.795 & 0.298 & 0.984 & 0.047 & 9.137\\
        
    \bottomrule
    \end{tabular}%
    }
    \caption{Complete results for the multimodal Sentiment Detection dataset. We observe the same trends as we do with the Crisis Humanitarianism dataset, demonstrating the generalizability of our approach.}
    \label{tab:app_results-sentiment}
\end{table*}

\subsection{Topical Coherence}
\label{sec:app_topic_model}
To measure the topical coherence between generated dilution and the original text, we compute the KL Divergence between the topic distributions of the two text segments --- i.e., $D_{KL} (P_{dilution} \vert \vert Q_{original})$. We train an LDA topic model~\cite{blei2003latent} using the text in a task-specific training set. The presented KL divergence scores are averaged over all the examples in the test set. We set the number of topics to be $20$ (based on topic coherence score) for the results presented in this paper. Additionally, we do not witness a change in the observed trends with variations in the chosen number of topics ($n \in \{5, 10, 15, 20\}$) for LDA topic modeling.

For implementing the Self-BLEU metric for quantifying diversity, we use NLTK's BLEU score function~\cite{loper2002nltk} and adopt the approach proposed in ~\citeauthor{zhu2018texygen} (\citeyear{zhu2018texygen}).

\subsection{Results on Sentiment Detection}
\label{sec:app_sentiment_results}
The main text presents an abridged version of the results on the Sentiment Detection dataset. The complete results are presented in Table \ref{tab:app_results-sentiment}.

\subsection{Human evaluation details}
\label{sec:app_human_evaluation}
For both our annotation tasks, we recruited annotators using Amazon Mechanical Turk. We set the criteria to `Master' annotators who had at least $90\%$ approval rate and were located in the United States. The rewards were set by assuming an hourly rate of 10 USD for all the annotators. In addition, the annotators were informed that the aggregate statistics of their annotations would be used and shared as part of academic research. 

The annotators were primed to identify real social media posts by showing them $5$ original multimodal examples. Previous research has demonstrated the role of providing examples in obtaining high-quality annotations~\cite{khashabi2021genie}.
For both our human evaluations, we also inserted some ``attention-check'' examples during the annotation tasks to ensure the annotators read the text carefully before responding. This was done by explicitly asking the annotators to mark a randomly-chosen score on the Likert scale regardless of the actual content. We discard the annotations from annotators who did not correctly respond to all the attention-check examples.

\begin{table}[!t]
    \centering
    \resizebox{\columnwidth}{!}{%
    \begin{tabular}{
    l  c c c c  c c c  c}
    \toprule
       & \multicolumn{4}{c}{\sc Classification Performance $\downarrow$} & \multicolumn{3}{c}{\sc Relevance $\uparrow$} & \textsc{Topical Diff. $\downarrow$}\\
      \cmidrule(lr){2-5}
      \cmidrule(lr){6-8}
      \cmidrule(lr){9-9}
        & $\mathbf{F_1}$ & $\mathbf{Prec.}$ & $\mathbf{Recall}$ & $\mathbf{Acc.}$ & 
        $\mathbf{Sim_{\bf \it text}}$ & $\mathbf{Sim_{\bf \it img}}$ & $\mathbf{Sim_{\bf \it corr}}$ &  \textbf{KL Div.}\\
        \midrule
        \textbf{XMD (Plain)}  & 0.663 & 0.654  & 0.669  & 0.671 & 0.586  & 0.237 & 0.979 & 11.012 \\
        \textbf{XMD (Adv)}  & 0.652  & 0.644  & 0.651  & 0.655 & 0.571  & 0.232 & 0.964 &  11.157 \\
        \textbf{XMD (Full)}  & 0.614 & 0.617 & 0.626 & 0.633 & 0.795 & 0.298 & 0.984 & 9.137\\
        
    \bottomrule
    \end{tabular}}
    \caption{Ablation results for the multimodal Sentiment Detection dataset.}
    \label{tab:results-sentiment-ablation}
\end{table}

\subsection{Ablations for Sentiment Detection}
The ablation results on the Sentiment Detection dataset are presented in Table \ref{tab:results-sentiment-ablation}. The results follow the same trends as discussed in Section \ref{sec:analysis} for the Crisis Humanitarianism dataset. 
\label{sec:app_ablation_sentiment}

\begin{table}[!t]
    \centering
    \resizebox{\columnwidth}{!}{%
    \begin{tabular}{l r l c}
    \toprule
         & \textbf{\# Words (std. dev.)} & \textbf{ Control tech.} & \textbf{Updated} $\mathbf{F_1}$  \\
         \midrule
        \textbf{Original} & 12.12 \small{(3.94)} & -- & 0.734 \\
        \textbf{Rule-based} & & & \\
        \hspace{3mm} Random URL & -- & repeat 5 times & 0.693\\
        \hspace{3mm} Image KW &  + 3.18 \small{(1.97)} & repeat 5 times & 0.711\\
        \hspace{3mm} Text KW & + 2.76 \small{(0.51)} & repeat 8 times & 0.726\\
        \hspace{3mm} Text + Image KW & + 5.85 \small{(2.13)} & repeat 4 times & 0.681\\
        \hspace{3mm} Similar image's desc & + 11.72 \small{(3.59)} & repeat twice &\\\midrule
        \textbf{Model-based} & & & \\
        \hspace{3mm} GPT & + 20.85 \small{(8.64)} & no change & 0.684\\
        \hspace{3mm} GPT-FT & + 22.87 \small{(6.52)} & no change & 0.628\\
        \hspace{3mm} MM Captions & + 9.11 \small{(1.12)} & repeat twice & 0.657\\
        \hspace{3mm} XLAN Captions & + 8.63 \small{1.96} & repeat twice & 0.662\\\midrule
        \textbf{Proposed} & & & \\
        \hspace{3mm} XMD (Plain) & + 36.43 \small{(10.16)} & truncate text & 0.649\\
        \hspace{3mm} XMD (Adv) & + 39.46 \small{(11.34)} & truncate text & 0.632\\
        \hspace{3mm} XMD (Full) & + 37.97 \small{(13.62)} & truncate text & 0.571\\
        \bottomrule
    \end{tabular}%
    }
    \caption{Numbers of words inserted by the dilution methods and classification performance after controlling for the number of inserted words (all methods have \textasciitilde20 words after modifications).}
    \label{tab:words_inserted}
\end{table}

\subsection{Length of Dilutions}
\label{sec:app_length_analysis}
To examine whether the drop in performance is contingent on the number of words inserted for dilution, we first report the number of words inserted using each of these methods (see Table \ref{tab:words_inserted}). Then, we control for the number of words inserted by employing either repetition or truncation so that each method inserts a comparable number of  \textasciitilde 20 words for dilution. As shown in Table \ref{tab:words_inserted}, even with comparable number of inserted words, the trends observed in Section \ref{sec:results} persist. This reinforces that it is not merely the dilutions' length that precipitates the drop in classification performance but the sensitivity to the inserted content. 

\begin{figure}[!t]
    \centering
    \includegraphics[width = 0.65\linewidth]{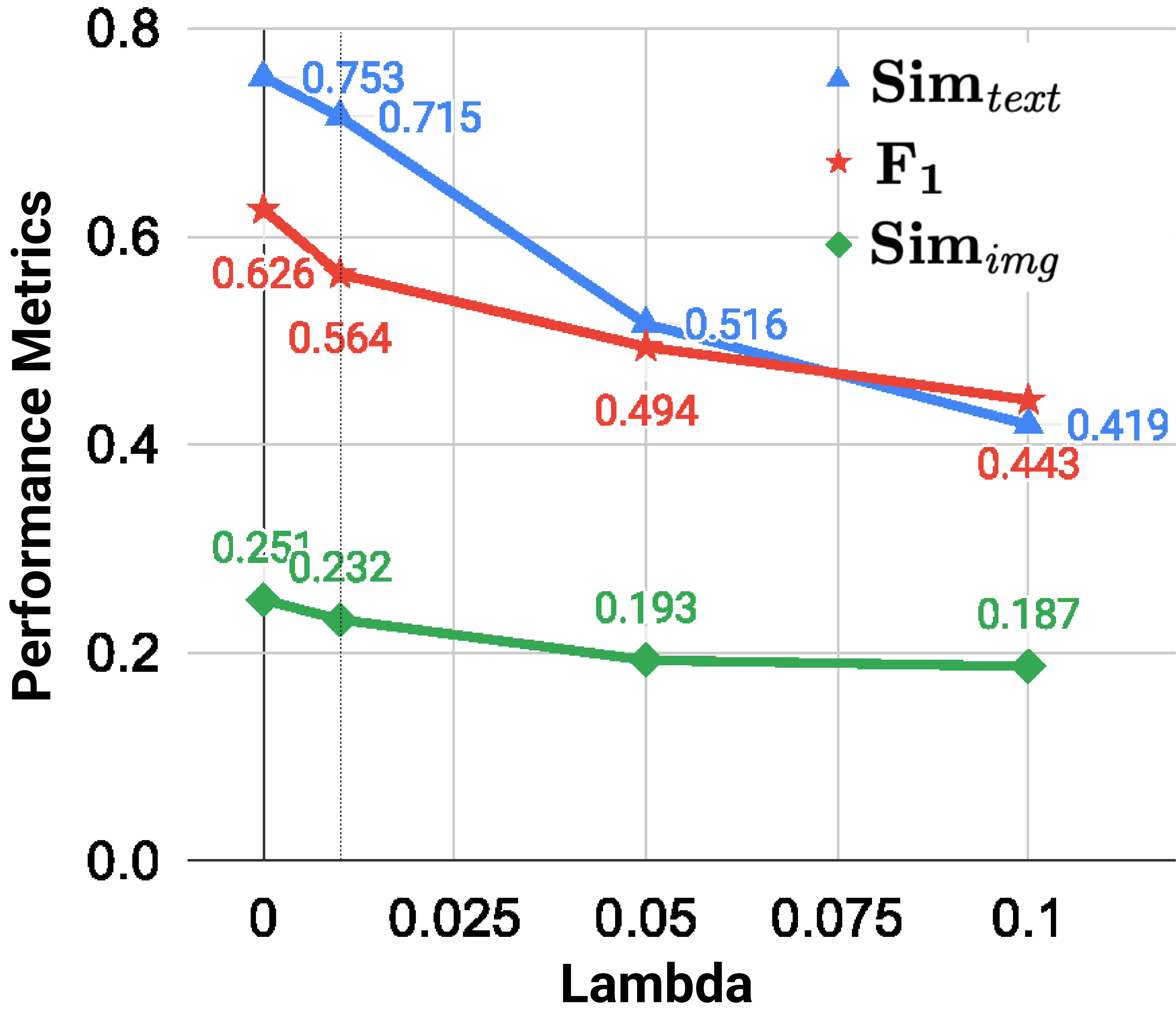}
    \caption{\textbf{Effect of varying $\mathbf{\lambda}$.} As $\lambda$ is increased, the adversarial effectiveness of the generated dilutions increases (lower $F_1$) but at the expense of relevance with original text \& image (lower \textbf{Sim}$_{text}$ and \textbf{Sim}$_{img}$). }
    \label{fig:lambda_variation}
\end{figure}

\subsection{Effect of variations in lambda}
\label{sec:app_lambda_variation}
Our main results and subsequent analyses are based on $\lambda = 0.01$, which controls the contribution of adversarial loss in the overall objective (see Equation \ref{eq:loss_formulation}). Figure \ref{fig:lambda_variation} shows the variation in classification performance on the crisis humanitarianism dataset with respect to the variations in $\lambda$. We find that as $\lambda$ increases, the classification performance deteriorates further. However, increasing $\lambda$ hurts the relevance of the generated dilution with the original text and image, as well as the topical coherence -- the relevance and coherence scores drop quickly as the relative contribution of $\mathcal{L}_{gen}$ is reduced. 

\subsection{Limitations}
\label{sec:app_limitations}

As indicated in Section \ref{sec:analysis}, in some scenarios, the extracted keywords from the images could be generic and do not extract meaningful keywords towards the specific task at hand. For instance, for multimodal fake news detection, the extracted keywords from pictures of celebrity faces are typically: \textit{man}, \textit{woman}, \textit{eye}, \textit{smile}, \textit{dress} etc. However, these keywords are unrelated to the larger (true/false) discourses centered around the celebrity. Similarly, for multimodal hate speech detection, the extracted keywords are often literal (such as \textit{hat}, \textit{clown}, \textit{monkey}) while the original text aims to establish provocative parallels like calling a person clown or associating certain groups with animals. Our current work is best applied to settings where the contextual relationship between the visual and textual modalities is straightforward, and extracted keywords provide a good representation of the cumulative expression. As part of our future work, we intend to develop cross-modal dilution strategies that can work with a wider variety of user-generated multimodal data. 

\end{document}